\documentclass[letterpaper, 10 pt, conference]{ieeeconf} 
\IEEEoverridecommandlockouts                              

\usepackage{todonotes} 
\usepackage{graphicx}
\usepackage{svg} 
\usepackage{textcomp}

\newcommand{\textapprox}{\raisebox{0.5ex}{\texttildelow}}
\title{\LARGE \bf
A Unified Framework for Real-Time Failure Handling in Robotics Using Vision-Language Models, Reactive Planner and Behavior Trees
}

\author{Faseeh Ahmad*$^{1}$, Hashim Ismail*$^{1}$, Jonathan Styrud$^{2, 3}$, Maj Stenmark$^{1}$ and Volker Krueger$^{1}$%
\thanks{*Equal Contribution}
\thanks{$^{1}$Lund University, Lund, Sweden. E-mail: {firstname.lastname}@cs.lth.se}%
\thanks{$^{2}$KTH Royal Institute of Technology, Stockholm, Sweden. E-mail: jstyrud@kth.se}%
\thanks{$^{3}$ABB Robotics, Västerås, Sweden}%
}

\begin{document}

    \maketitle
\thispagestyle{empty}
\pagestyle{empty}

\begin{abstract}
Robotic systems often face execution failures due to unexpected obstacles, sensor errors, or environmental changes. Traditional failure recovery methods rely on predefined strategies or human intervention, making them less adaptable. This paper presents a unified failure recovery framework that combines Vision-Language Models (VLMs), a reactive planner, and Behavior Trees (BTs) to enable real-time failure handling. Our approach includes pre-execution verification, which checks for potential failures before execution, and reactive failure handling, which detects and corrects failures during execution by verifying existing BT conditions, adding missing preconditions and, when necessary, generating new skills. The framework uses a scene graph for structured environmental perception and an execution history for continuous monitoring, enabling context-aware and adaptive failure handling. We evaluate our framework through real-world experiments with an ABB YuMi robot on tasks like peg insertion, object sorting, and drawer placement, as well as in AI2-THOR simulator. Compared to using pre-execution and reactive methods separately, our approach achieves higher task success rates and greater adaptability. Ablation studies highlight the importance of VLM-based reasoning, structured scene representation, and execution history tracking for effective failure recovery in robotics.
\end{abstract}

\section{Introduction}

Modern robotic systems excel in controlled environments, but struggle with dynamic environments such as small batch manufacturing, particularly in handling execution failures~\cite{LOFVING2018177}. Failures such as unexpected obstacles, sensor inaccuracies, or misaligned objects disrupt operations, causing costly delays~\cite{biomimetics9100612}. Unlike repetitive, pre-planned tasks in large-scale production, small batch manufacturing demands adaptability to frequent task variations. Similarly, in collaborative assembly lines, where robots work alongside humans, real time failure handling is crucial for safe and efficient execution~\cite{sharma2023adaptivecompliantrobotcontrol}. Developing autonomous failure recovery mechanisms that enable robots to detect, identify, and correct failures without human intervention is essential for improving reliability and reducing downtime~\cite{9561002}.

To address these challenges, failure recovery methods range from learning based approaches that rely on data driven policies to structured execution frameworks designed for modular and interpretable decision making. Many learning based methods employ end-to-end architectures where robotic control policies are trained directly from data~\cite{9945538,booher2024cimrlcombiningimitationreinforcement}. While effective across diverse tasks, these methods often lack interpretability and verifiability, making them unsuitable for safety critical domains requiring robust, failure resistant execution especially in high stakes environments where errors can damage expensive equipment or disrupt operations.

\begin{figure}[t]
    \centering
    \includegraphics[width=0.45\textwidth]{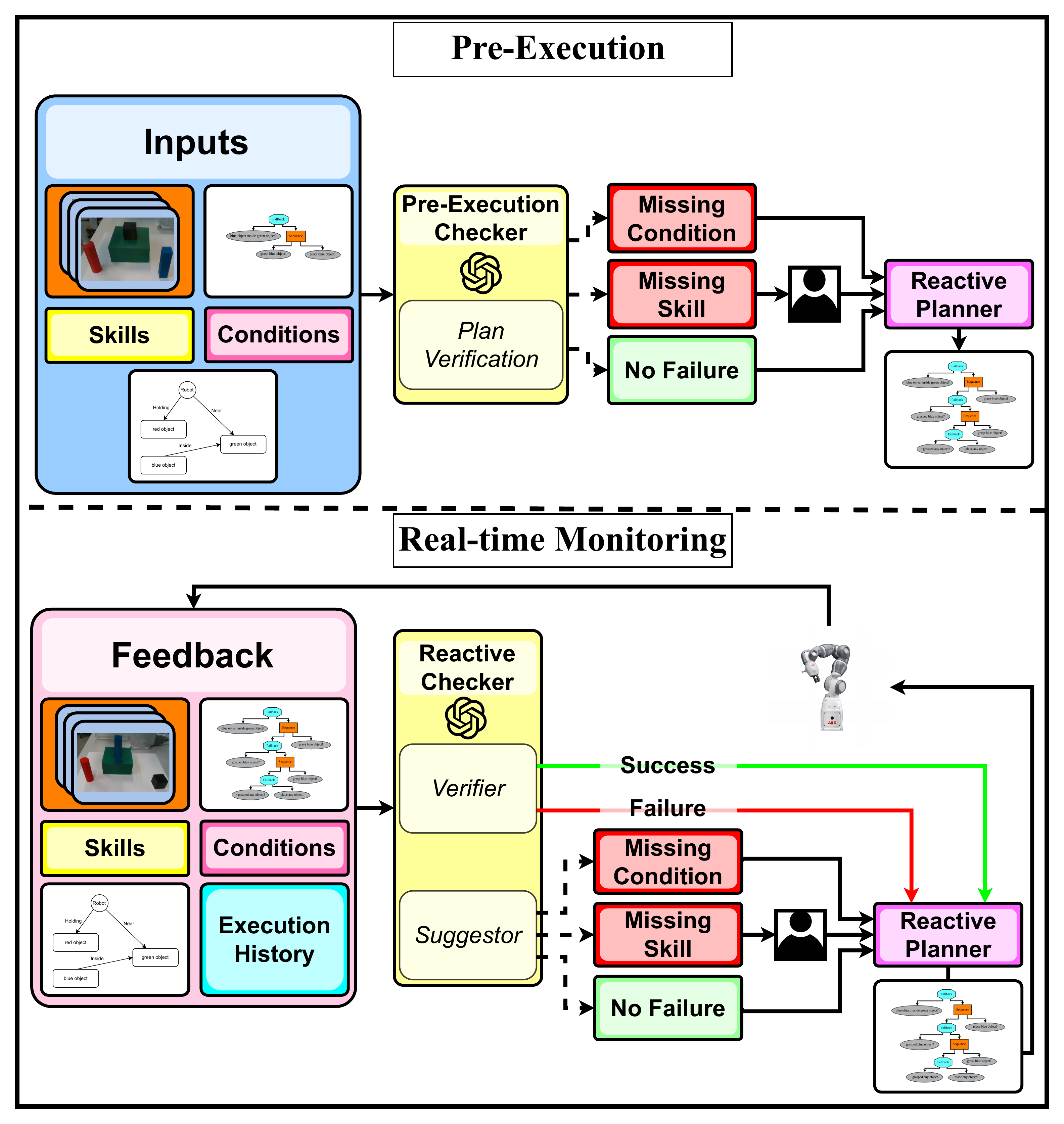}
    \caption{Overview of our approach, which consists of two phases: pre-execution verification and real-time monitoring. The pre-execution phase verifies the entire planned BT proactively using a VLM based on inputs (images, scene graphs, skills, and conditions). The real-time phase continuously monitors execution, where the VLM verifies preconditions, postconditions, and infers missing preconditions for individual skills using updated inputs and execution history. A reactive planner dynamically generates and adapts the BT as the robot’s execution policy.}
    \vspace{-0.5cm}
    \label{fig:approach}
\end{figure}

Structured execution frameworks, such as Behavior Trees (BTs)~\cite{colledanchise17ac}, provide a modular framework for verification, adaptation, and efficient failure recovery. They define execution policies as hierarchical compositions of reusable skills~\cite{ahmad2024adaptable}, enabling fine-grained monitoring while ensuring compliance with safety standards~\cite{biggar2022modularity}. Their modularity supports incremental recovery, avoiding the computational cost of full replanning~\cite{marzinotto142iicrai}. While BTs can be manually designed, reactive planners automate their generation using a backchaining approach that selects skills based on preconditions and postconditions~\cite{styrud2023bebop}. This allows robots to construct reactive execution policies that adapt to unexpected conditions in real-time without requiring full replanning.


In our prior work~\cite{ahmad2024addressing}, we introduced a failure recovery framework that used a Vision-Language Model (VLM) for pre-execution plan verification. The system analyzed input skills, execution conditions, the planned BT, and pre-execution images to assess whether the plan contained sufficient knowledge for successful execution. If critical preconditions or required skills were missing, it suggested modifications to prevent execution errors, reducing failures caused by incomplete task knowledge. However, this approach did not account for failures arising during execution due to unforeseen disturbances, environmental changes, or hardware errors.


While pre-execution verification helps prevent many failures, it cannot predict all possible execution-time issues. A robot may generate a valid pick-and-place plan, yet unexpected events, such as human intervention or object displacement, can still cause grasp failures. Addressing such failures requires real-time monitoring and corrective actions, which is only possible through a reactive mechanism. Without continuous failure monitoring, robots cannot effectively detect and adapt to failures as they occur, making reactive checks essential for robust autonomous execution.

Building on our prior work~\cite{ahmad2024addressing}, this paper presents a unified failure recovery framework that extends pre-execution plan verification with real-time execution monitoring (Figure~\ref{fig:approach}) to detect, identify, and correct errors dynamically. Our framework integrates reactive failure handling using a continuously updated execution history, which records skill execution states, timestamps, and scene graph updates for adaptive failure recovery. To improve situational awareness, we incorporate scene graphs that track object-object and robot-object spatial relationships throughout execution. Unlike~\cite{liu2023reflect}, which generates scene graphs post-execution, our method updates them continuously, enabling immediate detection of environmental changes. Additionally, while~\cite{ahmad2024addressing} suggested missing skills only pre-execution, our approach supports both pre-execution and reactive skill suggestions, ensuring failures are addressed proactively and dynamically. This work makes the following key contributions:

\begin{itemize} 
    \item A unified failure recovery framework integrating Vision-Language Models (VLMs), reactive planners, and Behavior Trees (BTs) for pre-execution failure verification and real-time reactive failure handling.
    \item Real-time failure detection, identification, and correction using an incrementally updated execution history that tracks skill conditions, execution timestamps, and scene graph updates.
    \item Experimental validation in AI2-THOR~\cite{ai2thor} and a real-world ABB YuMi robot, demonstrating improved failure recovery across diverse environments.
\end{itemize}

\section{Related Work}

Failure recovery in robotics has been extensively studied, from predefined strategies to modern learning-based techniques and Large Language Models (LLMs) for adaptive failure handling. This section reviews these methodologies and highlights the distinctions between existing works and our approach.

\subsection{Traditional Failure Recovery Strategies}

Early methods relied on human intervention, predefined recovery strategies, and automated solutions based on failure mode analysis. While human-in-the-loop strategies offer flexibility, they are labor-intensive and limit scalability~\cite{itadera2024motionpriorityoptimizationframework}. Predefined strategies handle known failure cases well but struggle with novel issues~\cite{wu2021automated}. Systematic failure analysis, such as Failure Mode and Effects Analysis (FMEA), requires expert knowledge and does not generalize to dynamic environments~\cite{jusuf2021review}. Automated recovery methods attempt autonomy but remain constrained by predefined failure modes~\cite{lei2023artificial, alves2020secure}. Unlike these approaches, our framework continuously updates a dynamic execution history for real-time failure detection and adaptation.

\subsection{Learning-Based Failure Recovery}

Recent approaches explore reinforcement learning (RL) and imitation learning (IL) to develop recovery strategies from experience~\cite{9945538, booher2024cimrlcombiningimitationreinforcement}. RL-based methods require extensive training in simulations, making real-world deployment difficult~\cite{duan2024ahavisionlanguagemodeldetectingreasoning}. IL-based methods like RACER~\cite{dai2024racerrichlanguageguidedfailure} improve recovery using demonstrations but struggle to generalize. Neuro-symbolic methods combine structured reasoning with learning, improving interpretability but facing scalability challenges~\cite{cornelio2024recoverneurosymbolicframeworkfailure, bougzime2025unlockingpotentialgenerativeai, zhang2024neurosymbolicaiexplainabilitychallenges}. Our approach avoids data-heavy training by leveraging Vision-Language Models (VLMs) for reasoning-based failure recovery, enabling flexible and context-aware corrections in real time.

\subsection{Failure Recovery with Large Language Models (LLMs) and Vision-Language Models (VLMs)}

LLMs and VLMs have become integral to robotic failure recovery due to their reasoning capabilities. Several approaches leverage LLM-based reasoning for failure detection and correction, including REFLECT~\cite{liu2023reflectsummarizingrobotexperiences}, AHA~\cite{duan2024ahavisionlanguagemodeldetectingreasoning}, DoReMi~\cite{guo2024doremigroundinglanguagemodel}, ReplanVLM~\cite{wang2024largelanguagemodelsrobotics}, RECOVER~\cite{cornelio2024recoverneurosymbolicframeworkfailure}, and Code-as-Monitor~\cite{zhou2024codeasmonitorconstraintawarevisualprogramming}. REFLECT provides hierarchical post-execution summaries but lacks real-time intervention. AHA fine-tunes a VLM for failure detection at task checkpoints but lacks structured execution policies. DoReMi enforces dynamic execution constraints but relies on LLM-generated constraints, introducing variability. ReplanVLM integrates pre-execution validation with execution monitoring using GPT-4V but depends on LLM-driven re-planning rather than structured failure handling.

Unlike these, our framework integrates a reactive planner and Behavior Trees (BTs) for structured, real-time failure handling at both pre-execution and reactive levels. \textit{RECOVER}\cite{cornelio2024recoverneurosymbolicframeworkfailure} uses ontology-driven neuro-symbolic reasoning for real-time failure detection but requires domain-specific engineering, limiting adaptability. \textit{Code-as-Monitor}\cite{zhou2024codeasmonitorconstraintawarevisualprogramming} translates natural language constraints into executable monitors for proactive (handling foreseeable failures) and reactive failure detection but lacks explicit recovery mechanisms. Unlike these, our execution history continuously updates skill execution states, enabling VLMs to analyze failures dynamically rather than post-execution. Compared to \textit{AHA} and \textit{ReplanVLM}, which focus on high-level reasoning or planning corrections, our approach ensures modular and adaptive failure recovery by integrating structured execution policies via BTs and a reactive planner. Additionally, recent work~\cite{chen2024integrating} explores intent-based BT planning using LLMs for goal interpretation, whereas our method actively modifies execution policies by suggesting missing preconditions, postconditions, and skills in real time, ensuring robust failure recovery in dynamic environments.

\section{Background}
In this section, we discuss the relevant concepts that serve as background knowledge for the paper.

\subsection{Behavior Trees}

Behavior Trees (BTs) are a hierarchical execution model valued for their modularity, flexibility, and reactivity in robotic decision-making~\cite{colledanchise2018behavior, iovino2020survey}. Originally developed for game AI, BTs now provide interpretable and scalable task execution in robotics~\cite{biggar2022modularity, colledanchise17ac}. Their structure simplifies behavior design, modification, and debugging while enabling real-time adaptation to dynamic environments~\cite{styrud2024automaticbehaviortreeexpansion}.

A BT is a directed acyclic graph where execution begins at the root node, propagating tick signals to evaluate and execute behaviors dynamically. Nodes return \textit{Success}, \textit{Failure}, or \textit{Running}, with control-flow nodes (e.g., \textit{Sequence}, \textit{Fallback}) managing execution order and execution nodes (e.g., \textit{action}, \textit{condition}) implementing robot skills. This structured execution enables task decomposition and fine-grained monitoring. Once adapted to handle a failure, the BT becomes a reusable execution policy, reducing reliance on model queries and improving efficiency over time.

\subsection{Reactive Planner}

Reactive planners generate Behavior Trees (BTs) dynamically using backchaining, selecting skills that satisfy goal conditions~\cite{8794128}. Starting from the goal, the planner works backward through skill preconditions and postconditions, iteratively expanding the BT until all conditions are met or a termination criterion is reached. This approach enables robots to adapt to environmental changes without full re-planning, leveraging BT modularity for flexible execution~\cite{styrud2023bebop}. The PDDL-based reactive planner used in this work follows~\cite{styrud2023bebop}, ensuring efficiency by removing redundant nodes and introducing composite subtrees for complex tasks. This facilitates real-time, autonomous failure recovery while maintaining computational efficiency. As backchaining inherently selects skills that achieve required postconditions, explicit VLM-generated postcondition suggestions are unnecessary.

\subsection{Vision-Language Models}
Vision-Language Models (VLMs) combine visual perception with language-based reasoning, making them effective for robotic failure recovery~\cite{cornelio2024recoverneurosymbolicframeworkfailure, dai2024racerrichlanguageguidedfailure}. They enable robots to detect, identify, and correct failures by analyzing execution conditions and task states.

In our prior work~\cite{ahmad2024addressing}, GPT-4o was used for pre-execution verification, where the VLM assessed if a planned execution contained sufficient knowledge to succeed. It performed three key tasks: failure detection (checking for potential failures based on available conditions), failure identification (diagnosing root causes by analyzing missing or incorrect preconditions), and failure correction (suggesting modifications such as adding missing preconditions or required skills). This approach reduced failures due to incomplete task knowledge but did not address execution-time failures from unforeseen disturbances or environmental changes.

This work extends VLMs to real-time execution monitoring and correction. The VLM continuously analyzes execution states, providing corrective suggestions based on evolving conditions. To improve reasoning, we integrate scene graphs that dynamically track object-object and robot-object relationships, improving failure detection. Additionally, an execution history records skill preconditions, postconditions, and execution timestamps, enabling structured failure analysis. By combining pre-execution checks with reactive real-time monitoring, our framework ensures continuous adaptation to failures, enhancing robustness in autonomous robotic execution.

\begin{figure*}[t] 
    \centering
    \includegraphics[width=1.0\textwidth]{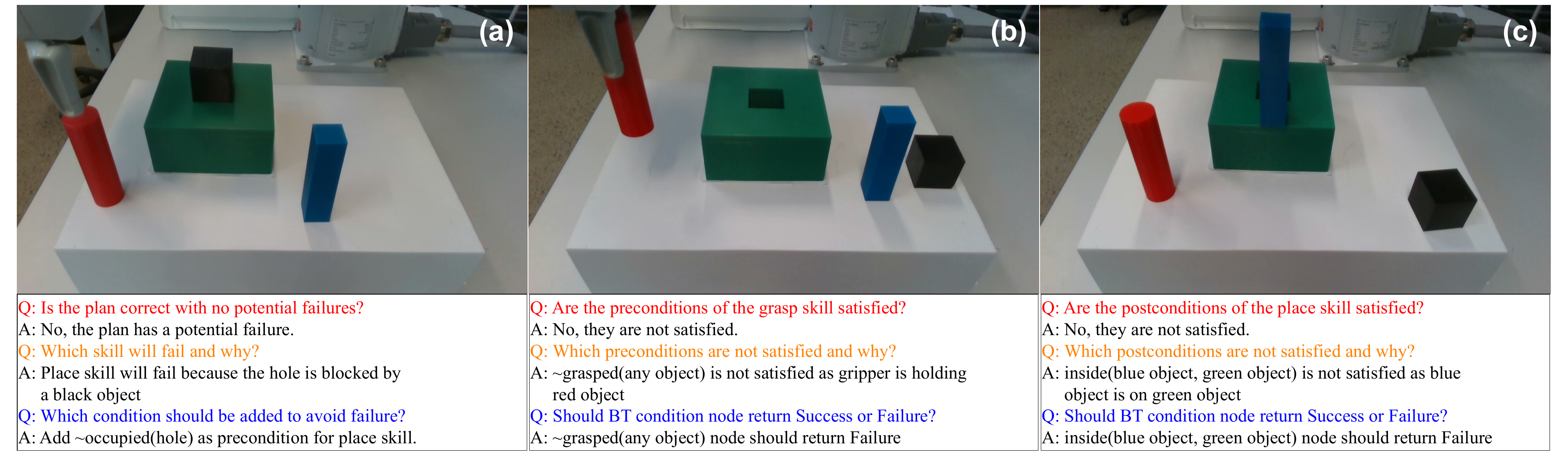} 
    \caption{Three failure instances with corresponding VLM responses. (a) Pre-execution verification detects that the black object blocks the hole, and the VLM suggests adding the missing precondition for the place skill. (b) Precondition verification identifies that the grasp skill fails due to an unmet condition, as the robot is already holding a red object. (c) Postcondition verification detects a failed placement since the blue object is on top of the green object instead of inside. Failure detection (red), identification (orange), and correction (blue) are indicated with corresponding VLM responses in black.}
    \vspace{-0.5cm}
    \label{fig:proactive_preverifier_postverifier}
\end{figure*}

\section{Approach}
To enable real-time robotic failure recovery, our framework integrates a reactive planner, Behavior Trees (BT), and Vision Language Models (VLM). The failure monitoring process is divided into pre-execution failure verification and real-time execution monitoring, each addressing failure detection, identification, and correction. Additionally, we extend the system with a scene graph and execution history to improve failure reasoning and adaptation. All failure handling mechanisms rely on the following key inputs:
\begin{itemize}
    \item \textbf{Images} capturing the scene from multiple angles using two cameras (front and side views) to improve spatial understanding.
    \item \textbf{Skills} with predefined pre- and postconditions.
    \item \textbf{Known conditions} for environment reasoning.
    \item \textbf{Scene graph} representing spatial object relations.
    \item \textbf{Behavior Tree (BT)} defining execution policy.
    \item \textbf{Execution history} (real-time only) tracking past actions and scene updates.
\end{itemize}

Failure handling follows a three-phase process: \textit{detection} identifies potential failures, \textit{identification} determines the root cause by pinpointing the affected skill and unmet condition, and \textit{correction} modifies the BT through precondition adjustments or skill additions to ensure successful execution. inspired by chain-of-thought~\cite{wei2022chain} reasoning, we structure failure recovery prompt into these three phases. This improves the VLM performance by guiding it step-by-step toward the correct solution. If no failure is detected during the detection phase, the system skips the identification and correction steps, optimizing computational efficiency in both pre-execution and real-time monitoring.
 
\begin{figure}[h] 
    \centering
    \includegraphics[width=0.4\textwidth]{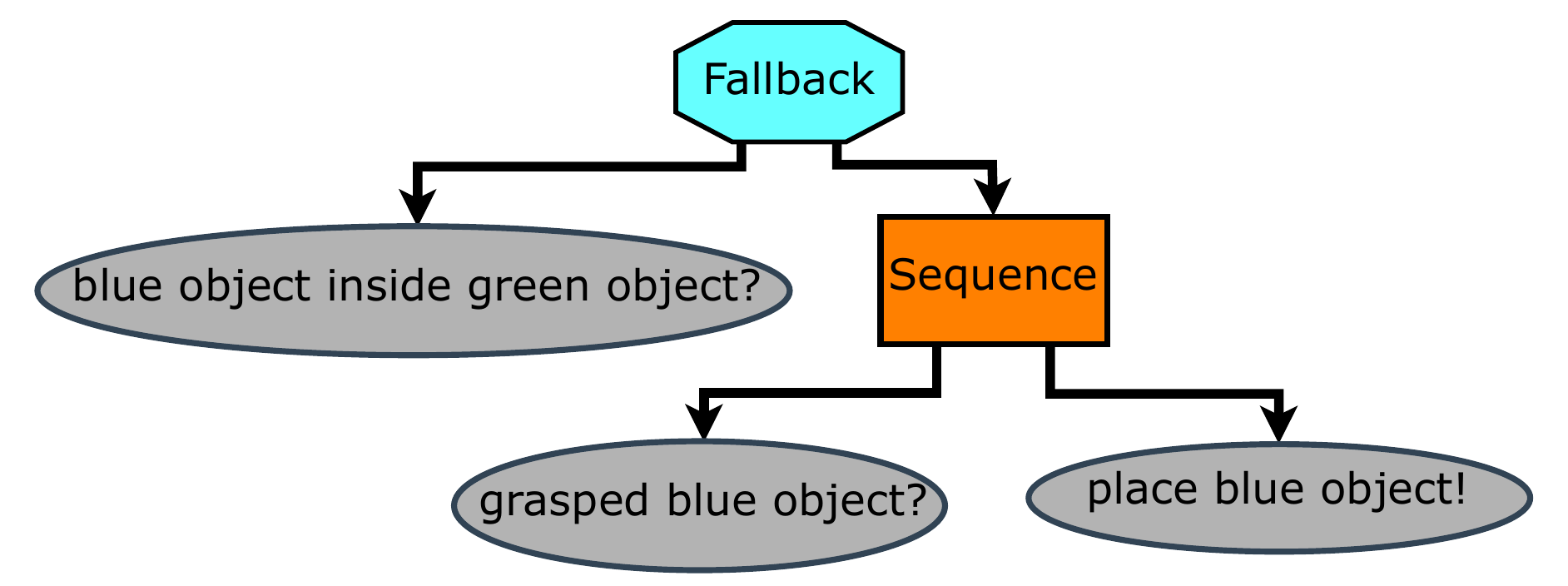} 
    \caption{BT of the peg-in-hole task without failure handling}
    \vspace{-0.25cm}
    \label{fig:peg_simple_BT}
\end{figure}
\begin{figure}[h] 
    \centering
    \includegraphics[width=0.4\textwidth]{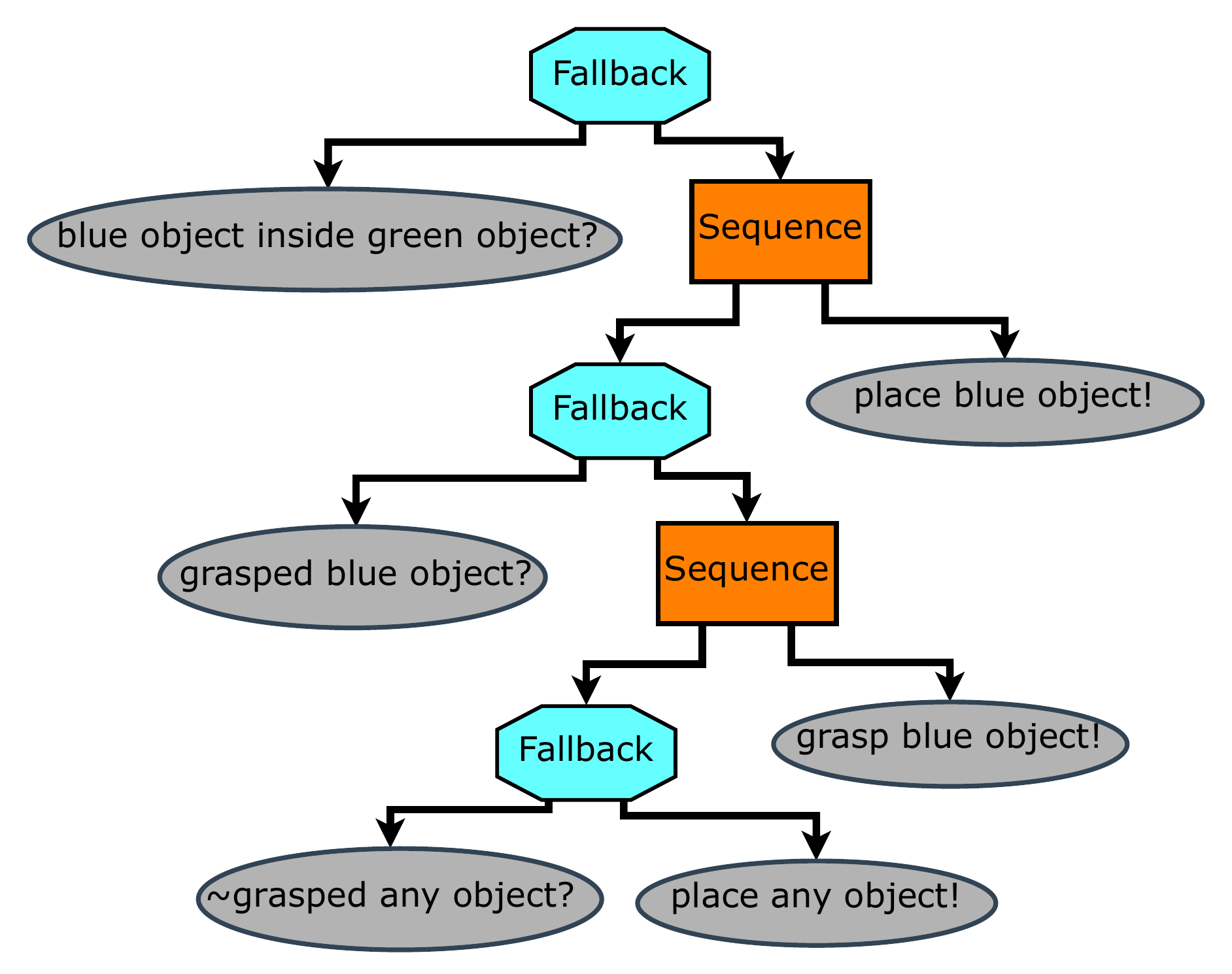} 
    \caption{Extended BT execution where a missing precondition is added, ensuring the gripper is empty before grasping target object.}
    \vspace{-0.5cm}
    \label{fig:peg_w_grasped_BT}
\end{figure}
\begin{figure*}[t] 
    \centering
    \includegraphics[width=1.0\textwidth]{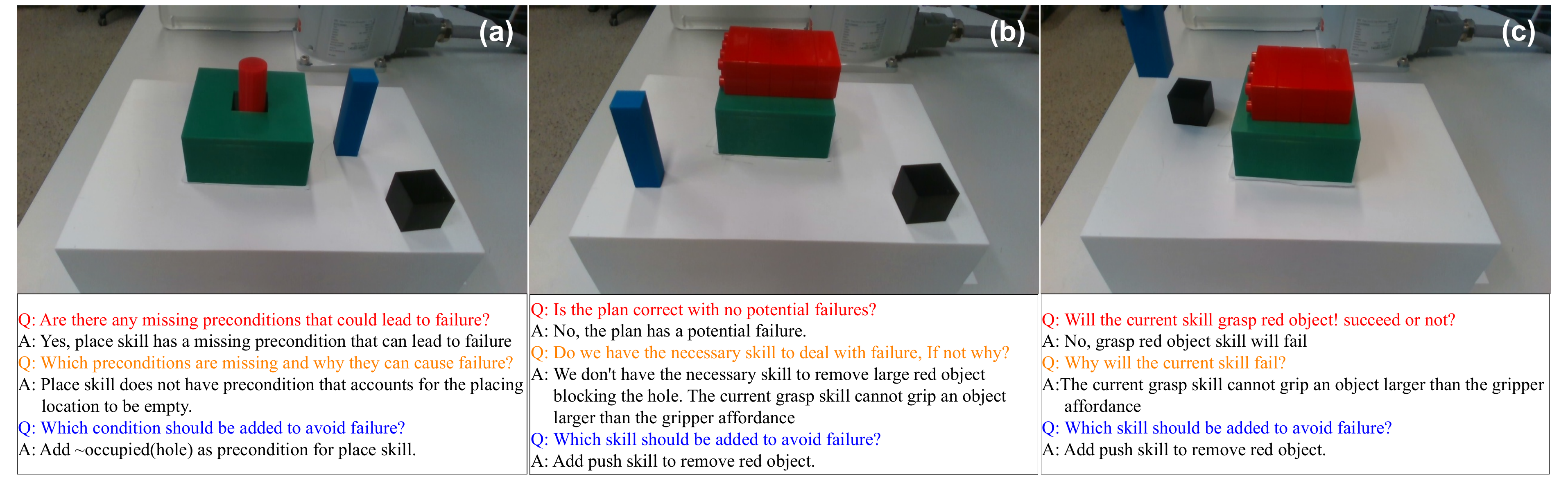} 
    \caption{The figure illustrates three failure scenarios and corresponding VLM responses. (a) Precondition suggestor: The red object inside the green object leads the VLM to identify a missing precondition for the place skill. (b) Pre-execution missing skill generation: The VLM identifies the need for a push skill to remove the red object. (c) Real-time missing skill generation: The VLM suggests generating the push skill during execution. Failure detection (red), identification (orange), and correction (blue) phases are depicted, with VLM responses in black.}
    \vspace{-0.5cm}
    \label{fig:presugg_proskill_reacskill}
\end{figure*}

To explain concretely our failure handling process, we use a peg-in-hole task, where the goal is to insert the blue object inside the green object, while red and black objects act as obstacles. Figures~\ref{fig:proactive_preverifier_postverifier} and~\ref{fig:presugg_proskill_reacskill} illustrate different failure types with VLM responses. These figures also show various prompts, color-coded to distinguish between failure detection, identification, and correction questions posed to the VLM~\footnote{Full prompts and code will be released after the submission process.}. From here onward, we will consider a BT for peg-in-hole task execution that does not yet account for failures, as shown in Figure~\ref{fig:peg_simple_BT}, unless specified otherwise.

\subsection{Pre-Execution Failure Verification}
Before execution~\cite{ahmad2024addressing}, we validate the planned BT by proactively checking for missing preconditions or potential execution failures. This step prevents errors before they occur, reducing failures caused by incomplete task knowledge. A GPT-4o-based VLM performs this verification by analyzing the inputs.

\begin{itemize}
    \item \textbf{Detection}: Flags anomalies where the planned BT may fail based on the current scene.  
    \textit{For example, in the peg-in-hole task, if a black cube blocks the hole, the pre-execution checker detects a potential failure (Figure~\ref{fig:proactive_preverifier_postverifier}(a)).}
    
    \item \textbf{Identification}: Pinpoints the failing skill and the root cause, whether due to missing knowledge or an incorrect assumption.  
    \textit{In this case, the VLM identifies that the \textit{place} skill will fail as the BT does not ensure the hole is unoccupied before placement (Figure~\ref{fig:proactive_preverifier_postverifier}(a)).}
    
    \item \textbf{Correction}: Suggests a missing precondition to update the BT and prevent failure.  
    \textit{Here, the system adds \textapprox{}occupied(hole) as a precondition for \textit{place}, prompting the reactive planner to remove the black cube before placement (Figure~\ref{fig:proactive_preverifier_postverifier}(a)).}
\end{itemize}

\subsection{Real-Time Failure Monitoring}  
While pre-execution verification minimizes failures, unexpected execution failures may still occur due to sensor inaccuracies, dynamic obstacles, or external disturbances. To handle these, we introduce a real-time failure monitoring module comprising a \textit{Verifier} and a \textit{Suggestor}. Both modules use the same inputs as pre-execution verification but incorporate continuously updated scene graphs, images, and execution history for improved reasoning.

\subsubsection{Verifier}
Ensures that execution aligns with expected conditions by performing \textit{precondition verification before execution} and \textit{postcondition verification after execution}.

\paragraph{Precondition Verification}
Before executing a skill, the \textit{Verifier} checks if the skill preconditions hold. Consider the BT in Figure~\ref{fig:peg_w_grasped_BT} with an existing  {\tt \small\textapprox{}grasped any object} precondition in this case.
\begin{itemize}
    \item \textbf{Detection}: Flags an anomaly if the preconditions for the skill in the BT are unmet.  
    \textit{For example, in the peg-in-hole task, if the robot has already grasped a red object but needs to grasp the blue object, the verifier detects an anomaly (Figure~\ref{fig:proactive_preverifier_postverifier}(b)).This failure can occur if a human intervenes after the pre-execution failure check by manually placing the red object inside the gripper.}

    \item \textbf{Identification}: Determines the violated precondition and the cause of failure.  
    \textit{In this case, it finds that the  {\tt \small\textapprox{}grasped any object}  precondition of the grasp skill is not satisfied (Figure~\ref{fig:proactive_preverifier_postverifier}(b)).}  

    \item \textbf{Correction}: Prevents execution by marking relevant preconditions as unsatisfied. The reactive planner will then automatically expand the BT to satisfy the marked preconditions.
    \textit{For instance, the BT adapts by placing the currently held object before attempting the new grasp (Figure~\ref{fig:proactive_preverifier_postverifier}(b)).}  
\end{itemize}

\paragraph{Postcondition Verification}
After executing a skill, the \textit{Verifier} checks if expected postconditions hold.  
\begin{itemize}
    \item \textbf{Detection}: Flags an anomaly if the executed skill fails to meet its postconditions.  
    \textit{For instance, if the robot places the blue object on top of the hole instead of inside, the verifier detects a failure (Figure~\ref{fig:proactive_preverifier_postverifier}(c)).}  

    \item \textbf{Identification}: Identifies the violated postcondition and the cause of failure.  
    \textit{Here, it finds that the “inside” condition is violated because the object is on top rather than inside (Figure~\ref{fig:proactive_preverifier_postverifier}(c)).}  

    \item \textbf{Correction}: Returns \textit{Failure}, triggering the reactive planner to adjust execution dynamically.  
    \textit{The BT reattempts placement in the next tick (Figure~\ref{fig:proactive_preverifier_postverifier}(c)).}  
\end{itemize}
\subsubsection{Suggestor}
The \textit{Suggestor} dynamically infers missing preconditions when a skill fails due to unmet conditions.

\begin{itemize}
    \item \textbf{Detection}: Flags an anomaly when a skill is likely to fail due to an unmet precondition.  
    \textit{For example, in the peg-in-hole task, the red object is already occupying the hole (Figure~\ref{fig:presugg_proskill_reacskill}(a)).}

    \item \textbf{Identification}: Identifies the missing precondition and the cause of failure.  
    \textit{In this case, it determines that the place skill is missing a precondition ensuring the hole is empty before insertion (Figure~\ref{fig:presugg_proskill_reacskill}(a)).}

    \item \textbf{Correction}: Suggests the missing precondition, prompting the BT to update accordingly.  
    \textit{The model suggests \textapprox{}occupied(hole) as a precondition, allowing the reactive planner to expand the BT accordingly (Figure~\ref{fig:presugg_proskill_reacskill}(a)).}
\end{itemize}
\subsection{Skill Addition}
While modifying preconditions can resolve many failures, some cases require introducing new skills. The skill addition can be suggested either pre-execution or reactively depending on when the potential failure case arises. The pre-execution stage implements our prior work in the \cite{ahmad2024addressing} paper. if no existing skill can address a detected failure,  the system suggests a missing skill (Figure \ref{fig:presugg_proskill_reacskill}(b)). In the reactive phase,  the VLM checks execution feasibility before executing every skill. If the current skill is predicted to fail, a missing skill is suggested to remove the failure (Figure \ref{fig:presugg_proskill_reacskill}(c)).
\begin{itemize}
    \item \textbf{Detection}: Identifies when the available skills can not resolve a failure.  
    \textit{For example, in the peg-in-hole task, if a non-pickable object blocks the hole and the system detects an unresolved failure (Figure~\ref{fig:presugg_proskill_reacskill}(b)).}
    
    \item \textbf{Identification}: Determines the missing capability and the skill that fails due to this limitation.  
    \textit{In this case, the pick skill fails because the object is non-pickable (Figure~\ref{fig:presugg_proskill_reacskill}(b)).}
    
    \item \textbf{Correction}: The VLM suggests a new skill to resolve the failure, ensuring compatibility with the robot’s world model. The suggestion includes:  
    \begin{itemize}
        \item The name of the missing skill.
        \item A code template defining the skill.
        \item Predefined preconditions and postconditions.
    \end{itemize}
    
    \textit{For example, if a robot cannot grasp an object, the VLM may suggest a “Push” skill as an alternative, providing a skill description with predefined conditions (Figure~\ref{fig:presugg_proskill_reacskill}(b)).} Figure~\ref{fig:presugg_proskill_reacskill}(c) illustrates the reactive version occurring during execution, where the robot first places the blue object on the table before executing the "Push" skill to move the red object. To ensure consistency, the system restricts the VLM to known world model conditions, preventing arbitrary condition generation.
\end{itemize}

\subsection{Scene Graph Representation}
To enable real-time monitoring, our system maintains an evolving scene graph that tracks spatial relationships between objects and the robot. Unlike REFLECT~\cite{liu2023reflect}, which regenerates the scene graph from scratch at each timestep, our approach continuously updates it by modifying relationships and adding or removing nodes as needed.

The scene graph is constructed using:
\begin{itemize}
    \item \textbf{RGB-D images and point clouds} for capturing scene depth and object positioning.
    \item \textbf{Grounding DINO}~\cite{liu2024groundingdinomarryingdino} for object detection and \textbf{SAM2}~\cite{ravi2024sam2segmentimages} for instance segmentation and tracking.
    \item \textbf{RANSAC and PCA-based pose estimation} to estimate 6D object poses.
\end{itemize}

Continuous updates improve efficiency and ensure execution consistency. \textit{For example, in the peg-in-hole task, when the robot inserts the blue object into the green one, our system updates the scene graph by modifying the "on" relation to "inside" without reconstructing the entire graph.}

\subsection{Execution History}
The execution history maintains a log of skill executions, condition verification results, and environmental changes. Instead of explicit failure logging, which assumes perfect execution state knowledge, our approach captures execution traces via changes in the scene graph that help infer failures and inconsistencies.

\begin{itemize}
    \item \textbf{Skill execution records}: Logs executed skills with timestamps.
    \item \textbf{Precondition and postcondition verification}: Tracks whether preconditions were met before execution and if postconditions held afterward.
    \item \textbf{Scene graph updates}: Records object positions and relationships before and after execution to analyze deviations.
\end{itemize}

\textit{For example, in the peg-in-hole task, if the blue peg is placed on top of the green hole instead of inside, the execution history logs the "Place" skill execution with its timestamp. The system records that the precondition was satisfied (e.g., the peg was grasped), but postcondition verification fails as the peg's spatial relation does not match the expected "inside" condition. The scene graph update reflects this deviation, showing the peg as "on" rather than "inside" the hole.}

This structured history enables real-time adaptation by detecting execution anomalies, allowing the system to refine failure handling based on observed task progression.

\section{Experiments and Results}

We evaluate our failure recovery framework through both simulation benchmarks and real-world experiments. In simulation, we use benchmark tasks from REFLECT~\cite{liu2023reflect} in AI2-THOR~\cite{ai2thor}, assessing how our system adapts to predefined failure cases. For real-world validation, we implement our framework on a robotic platform to evaluate its effectiveness in handling failures in physical environments.

\subsection{Simulation Experiments}
We evaluate our framework on REFLECT benchmark tasks, where failures occur during execution and are corrected post-execution using hierarchical summaries and scene graphs~\cite{liu2023reflectsummarizingrobotexperiences}. However, REFLECT lacks real-time adaptation, as failures are only detected and corrected after task completion.

Our approach instead uses a reactive planner and BTs to dynamically generate execution policies, enabling real-time monitoring and immediate failure correction. Unlike REFLECT, which reconstructs a new scene graph per execution, our system continuously updates it. Additionally, while REFLECT relies on LLM-generated post-execution corrections without correctness guarantees, our reactive planner ensures correctness through structured preconditions and postconditions.

We successfully applied our framework to all REFLECT benchmark tasks, achieving a 100\% success rate across multiple runs. Real-time monitoring was sufficient, making pre-execution checks unnecessary. The \textit{Verifier} ensured execution correctness, while the \textit{Suggestor} resolved missing preconditions. Unlike REFLECT, which evaluates explanation, localization, and replanning success, we assess overall task completion. Since failures are proactively verified and reactively corrected during execution, post-execution replanning is unnecessary, reducing reliance on retrospective reasoning.

Key differences between REFLECT and our approach are summarized in Table~\ref{tab:reflect_comparison}.

\begin{table}[h]
\centering
\small 
\renewcommand{\arraystretch}{1.0} 
\setlength{\tabcolsep}{4pt} 
\caption{Qualitative Comparison of REFLECT and Our Approach}
\label{tab:reflect_comparison}
\begin{tabular}{|c|c|c|}
\hline
\textbf{Feature} & \textbf{REFLECT} & \textbf{Our Approach} \\ 
\hline
Execution Plan & Manually designed & Reactive BT \\ 
\hline
Failure Handling & Post-execution & Real-time \\ 
\hline
\begin{tabular}[c]{@{}c@{}}Scene Graph\\ Update\end{tabular} & \begin{tabular}[c]{@{}c@{}}Reconstructed\\ post-execution\end{tabular} & \begin{tabular}[c]{@{}c@{}}Maintained\\ incrementally\end{tabular} \\ 
\hline
Failure Detection & Post-execution & Real-time \\ 
\hline
Plan Correction & LLM-generated & Reactive BT \\ 
\hline
\end{tabular}
\vspace{-0.5cm}
\end{table}

\begin{table}[t]
\centering
\small 
\renewcommand{\arraystretch}{0.9} 
\setlength{\tabcolsep}{3pt} 
\caption{Comparison of Failure Recovery Baselines}
\label{tab:baseline_comparison}
\begin{tabular}{|c|c|c|c|}
\hline
\textbf{Metric} & \textbf{Pre-execution} & \textbf{Reactive} & \textbf{\begin{tabular}[c]{@{}c@{}}Pre-execution +\\ Reactive (Ours)\end{tabular}} \\ \hline
\textbf{\begin{tabular}[c]{@{}c@{}}Task Success \\ Rate\end{tabular}}           & 31.25\% & 100\% & \textbf{100\%} \\ \hline
\textbf{\begin{tabular}[c]{@{}c@{}}Failure Detection \\ Rate\end{tabular}}      & 31.25\% & 100\% & \textbf{100\%} \\ \hline
\textbf{\begin{tabular}[c]{@{}c@{}}Failure Identification \\ Rate\end{tabular}} & 100\% & 100\% & \textbf{100\%} \\ \hline
\textbf{\begin{tabular}[c]{@{}c@{}}Correction Success \\ Rate\end{tabular}}     & 100\% & 100\% & \textbf{100\%} \\ \hline
\textbf{\begin{tabular}[c]{@{}c@{}}Skill Suggestion \\ Accuracy\end{tabular}}   & 50\% & 100\% & \textbf{100\%} \\ \hline
\end{tabular}
\vspace{-0.5cm}
\end{table}

\subsection{Real-World Experiments}
For real-world validation, we deployed our framework on an ABB YuMi robot equipped with an RGB-D camera. We assessed its failure recovery capabilities across three tasks:

\begin{itemize}
    \item \textbf{Peg-in-hole}: Inserting a peg into a hole with varying initial placements.
    \item \textbf{Object Sorting}: Sorting objects by color into designated locations.
    \item \textbf{Drawer Placement}: Placing an object inside a drawer.
\end{itemize}

Failures were introduced by modifying object placements, adding obstructions, or altering task constraints. Additionally, human intervention was used to induce failures during execution.

\subsubsection{Baseline Approaches}
We compared our approach against two baselines to assess the benefits of integrating pre-execution and reactive failure recovery mechanisms:

\begin{itemize}
    \item \textbf{Pre-execution}: Check for plan verification~\cite{ahmad2024addressing}.
    \item \textbf{Reactive}: Detect and correct failures during execution.
    \item \textbf{Our Approach (Pre-execution + Reactive)}: Combine pre-execution validation and real-time monitoring to prevent and correct failures dynamically.
\end{itemize}

Table~\ref{tab:baseline_comparison} provides a quantitative comparison, showing that \textit{Pre-execution + Reactive} achieves the highest performance. While \textit{Reactive} matches its failure handling, it is significantly more expensive due to increased VLM queries, additional skill executions, and longer execution times. For instance, without pre-execution checks, a robot may start execution only to realize mid-task that a required object is missing, forcing backtracking and reactive correction causing delays and inefficiencies~\footnote{See the video submission for details.}. In contrast, our approach resolves pre-execution failures proactively whenever possible, reducing computational and execution overhead. Meanwhile, \textit{Pre-execution} has lower accuracy as it addresses failures only before execution but remains the most efficient, avoiding costly real-time interventions. This highlights the trade-off between execution success and efficiency, where pre-execution handling is computationally cheaper but insufficient for real-time failures.

\subsubsection{Evaluation Metrics}
We evaluated our framework's ability to detect, identify, and correct failures across 16 pre-recorded failure cases, repeating each experiment 10 times. To assess false positives, we also ran each task 10 times without introducing failures.

We measured the system’s accuracy in detecting failures, correctly identifying their root causes, and successfully correcting them. Additionally, we analyzed the proportion of pre-execution failures handled versus those requiring real-time intervention and assessed the accuracy of skill suggestions.

Table~\ref{tab:baseline_comparison} summarizes the performance across these metrics. Our framework achieved a perfect 100\% accuracy on all tasks, demonstrating strong failure recognition and reasoning capabilities. No false positives occurred when running the tasks without failures. Given that failure recovery systems are designed to achieve near-perfect accuracy, these results align with expectations. Future work should focus on evaluating the framework on more complex benchmarks to further assess its scalability and robustness.

\subsubsection{Ablation Studies and Summary of Findings}
To evaluate key components, we conducted ablation studies by selectively removing elements and analyzing their impact on failure recovery.

\begin{itemize}
    \item \textbf{VLM vs. LLM}: Removing vision input weakens spatial and scene-aware failure detection, limiting object relation reasoning. Success drops from 100\% (2 images) to 98\% (1 image) and 95\% (no images), assuming scene graph accuracy, which is not always guaranteed.
    
    \item \textbf{Scene Graph Contribution}: Assists spatial reasoning and removes scene ambiguity. Without it, success drops to 91.25\%, highlighting its role in structured failure prediction.
    
    \item \textbf{Execution History Effectiveness}: Omitting execution history tracking did not significantly impact results, as we observed similar success rates with and without it. However, this does not imply that execution history is ineffective; its benefits may become more evident in more complex benchmarks.
\end{itemize}

Our findings confirm that combining pre-execution and reactive failure handling improves task success. Pre-execution checks prevent plan failures, while real-time monitoring improves adaptability. VLM-based reasoning strengthens failure detection and correction, and scene graphs with execution tracking improve system reliability by maintaining structured environmental context. These results validate our framework’s effectiveness in autonomous failure recovery across diverse robotic tasks. 

\section{Conclusion and Future Work}
This paper presented a unified failure recovery framework integrating VLMs, a reactive planner, and Behavior Trees (BTs) for pre-execution failure detection and reactive recovery in robotic execution. By incorporating a scene graph for structured perception and execution history for real-time monitoring, our approach dynamically adapts to failures, minimizing execution disruptions. Experimental validation on an ABB YuMi robot and simulation benchmarks demonstrated its superiority over using pre-execution and reactive methods separately. Ablation studies confirmed the importance of VLMs, structured scene understanding, and execution summaries in enhancing system reliability.

In the future, we aim to enhance our framework by integrating video and audio inputs for improved context-aware task monitoring. We plan to fine-tune open-source multi-modal models for failure handling, reducing computational costs and improving efficiency. Additionally, we will leverage Vision-Language Action (VLA) models for autonomous skill generation with structured preconditions and postconditions, ensuring quality through static and integration checks. To extend real-time monitoring, we will incorporate holding conditions for proactive failure checking during execution. These advancements will enhance autonomous failure recovery, making robotic systems more adaptable and self-sufficient.

\section{Acknowledgements}
We thank Jialong Li for valuable discussions. This work was supported by the Wallenberg AI, Autonomous Systems, and Software Program (WASP) through the Knut and Alice Wallenberg Foundation and by Vinnova (NextG2Com, ref. no. 2023-00541). Experiments were partly conducted at ABB Corporate Research Center, Västerås, Sweden, with financial support from WASP. Generative AI tools were used for editing, including grammar and sentence structuring.

\addtolength{\textheight}{-12cm}   








\bibliography{2025-ROS}

\begin{thebibliography}{10}
\providecommand{\url}[1]{#1}
\csname url@rmstyle\endcsname
\providecommand{\newblock}{\relax}
\providecommand{\bibinfo}[2]{#2}
\providecommand\BIBentrySTDinterwordspacing{\spaceskip=0pt\relax}
\providecommand\BIBentryALTinterwordstretchfactor{4}
\providecommand\BIBentryALTinterwordspacing{\spaceskip=\fontdimen2\font plus
\BIBentryALTinterwordstretchfactor\fontdimen3\font minus \fontdimen4\font\relax}
\providecommand\BIBforeignlanguage[2]{{%
\expandafter\ifx\csname l@#1\endcsname\relax
\typeout{** WARNING: IEEEtran.bst: No hyphenation pattern has been}%
\typeout{** loaded for the language `#1'. Using the pattern for}%
\typeout{** the default language instead.}%
\else
\language=\csname l@#1\endcsname
\fi
#2}}

\bibitem{LOFVING2018177}
\BIBentryALTinterwordspacing
M.~Löfving, P.~Almström, C.~Jarebrant, B.~Wadman, and M.~Widfeldt, ``Evaluation of flexible automation for small batch production,'' \emph{Procedia Manufacturing}, vol.~25, pp. 177--184, 2018, proceedings of the 8th Swedish Production Symposium (SPS 2018). [Online]. Available: \url{https://www.sciencedirect.com/science/article/pii/S2351978918305912}
\BIBentrySTDinterwordspacing

\bibitem{biomimetics9100612}
\BIBentryALTinterwordspacing
R.~Liu, G.~Wan, M.~Jiang, H.~Chen, and P.~Zeng, ``Autonomous robot task execution in flexible manufacturing: Integrating pddl and behavior trees in ariac 2023,'' \emph{Biomimetics}, vol.~9, no.~10, 2024. [Online]. Available: \url{https://www.mdpi.com/2313-7673/9/10/612}
\BIBentrySTDinterwordspacing

\bibitem{sharma2023adaptivecompliantrobotcontrol}
\BIBentryALTinterwordspacing
E.~Sharma, C.~Henke, A.~Mitrevski, and P.~G. Plöger, ``Adaptive compliant robot control with failure recovery for object press-fitting,'' 2023. [Online]. Available: \url{https://arxiv.org/abs/2307.08274}
\BIBentrySTDinterwordspacing

\bibitem{9561002}
R.~Wu, S.~Kortik, and C.~H. Santos, ``Automated behavior tree error recovery framework for robotic systems,'' in \emph{2021 IEEE International Conference on Robotics and Automation (ICRA)}, 2021, pp. 6898--6904.

\bibitem{9945538}
S.~Kobayashi and T.~Shibuya, ``Reinforcement learning to efficiently recover control performance of robots using imitation learning after failure,'' in \emph{2022 IEEE International Conference on Systems, Man, and Cybernetics (SMC)}, 2022, pp. 1147--1154.

\bibitem{booher2024cimrlcombiningimitationreinforcement}
\BIBentryALTinterwordspacing
J.~Booher, K.~Rohanimanesh, J.~Xu, V.~Isenbaev, A.~Balakrishna, I.~Gupta, W.~Liu, and A.~Petiushko, ``Cimrl: Combining imitation and reinforcement learning for safe autonomous driving,'' 2024. [Online]. Available: \url{https://arxiv.org/abs/2406.08878}
\BIBentrySTDinterwordspacing

\bibitem{colledanchise17ac}
M.~Colledanchise and P.~Ögren, \emph{Behavior {{Trees}} in {{Robotics}} and {{AI}}: {{An Introduction}}}.\hskip 1em plus 0.5em minus 0.4em\relax Chapman \& Hall/CRC Press, 2017.

\bibitem{ahmad2024adaptable}
F.~Ahmad, M.~Mayr, S.~Suresh-Fazeela, and V.~Krueger, ``Adaptable recovery behaviors in robotics: A behavior trees and motion generators (btmg) approach for failure management,'' in \emph{2024 IEEE 20th International Conference on Automation Science and Engineering (CASE)}.\hskip 1em plus 0.5em minus 0.4em\relax IEEE, 2024, pp. 1815--1822.

\bibitem{biggar2022modularity}
O.~Biggar, M.~Zamani, and I.~Shames, ``On modularity in reactive control architectures, with an application to formal verification,'' \emph{ACM Transactions on Cyber-Physical Systems (TCPS)}, vol.~6, no.~2, pp. 1--36, 2022.

\bibitem{marzinotto142iicrai}
A.~Marzinotto, M.~Colledanchise, C.~Smith, and P.~Ögren, ``Towards a unified behavior trees framework for robot control,'' in \emph{2014 {{IEEE International Conference}} on {{Robotics}} and {{Automation}} ({{ICRA}})}, 2014, pp. 5420--5427.

\bibitem{styrud2023bebop}
J.~Styrud, M.~Mayr, E.~Hellsten, V.~Krueger, and C.~Smith, ``Bebop--combining reactive planning and bayesian optimization to solve robotic manipulation tasks,'' in \emph{2024 International Conference on Robotics and Automation ({{ICRA}})}.\hskip 1em plus 0.5em minus 0.4em\relax {IEEE}, 2024.

\bibitem{ahmad2024addressing}
F.~Ahmad, J.~Styrud, and V.~Krueger, ``Addressing failures in robotics using vision-based language models (vlms) and behavior trees (bts),'' \emph{arXiv preprint arXiv:2411.01568}, 2024, accepted at European Robotics Forum (ERF) 2025.

\bibitem{liu2023reflect}
Z.~Liu, A.~Bahety, and S.~Song, ``Reflect: Summarizing robot experiences for failure explanation and correction,'' \emph{arXiv preprint arXiv:2306.15724}, 2023.

\bibitem{ai2thor}
E.~Kolve, R.~Mottaghi, W.~Han, E.~VanderBilt, L.~Weihs, A.~Herrasti, D.~Gordon, Y.~Zhu, A.~Gupta, and A.~Farhadi, ``{AI2-THOR: An Interactive 3D Environment for Visual AI},'' \emph{arXiv}, 2017.

\bibitem{itadera2024motionpriorityoptimizationframework}
\BIBentryALTinterwordspacing
S.~Itadera and Y.~Domae, ``Motion priority optimization framework towards automated and teleoperated robot cooperation in industrial recovery scenarios,'' 2024. [Online]. Available: \url{https://arxiv.org/abs/2308.15044}
\BIBentrySTDinterwordspacing

\bibitem{wu2021automated}
R.~Wu, S.~Kortik, and C.~H. Santos, ``Automated behavior tree error recovery framework for robotic systems,'' in \emph{2021 IEEE International Conference on Robotics and Automation (ICRA)}.\hskip 1em plus 0.5em minus 0.4em\relax IEEE, 2021, pp. 6898--6904.

\bibitem{jusuf2021review}
F.~Jusuf, A.~Susanto, A.~Waluyo, and N.~Siwhan, ``Review on defenses against common cause failures on digital safety system,'' in \emph{AIP Conference Proceedings}, vol. 2374, no.~1.\hskip 1em plus 0.5em minus 0.4em\relax AIP Publishing, 2021.

\bibitem{lei2023artificial}
Y.~Lei, J.~Wilch, B.~Rupprecht, and B.~Vogel-Heuser, ``Artificial intelligence planning of failure recovery strategies in discrete manufacturing automation,'' in \emph{2023 IEEE 19th International Conference on Automation Science and Engineering (CASE)}.\hskip 1em plus 0.5em minus 0.4em\relax IEEE, 2023, pp. 1--8.

\bibitem{alves2020secure}
L.~V. Alves and P.~N. Pena, ``Secure recovery procedure for manufacturing systems using synchronizing automata and supervisory control theory,'' \emph{IEEE Transactions on Automation Science and Engineering}, vol.~19, no.~1, pp. 486--496, 2020.

\bibitem{duan2024ahavisionlanguagemodeldetectingreasoning}
\BIBentryALTinterwordspacing
J.~Duan, W.~Pumacay, N.~Kumar, Y.~R. Wang, S.~Tian, W.~Yuan, R.~Krishna, D.~Fox, A.~Mandlekar, and Y.~Guo, ``Aha: A vision-language-model for detecting and reasoning over failures in robotic manipulation,'' 2024. [Online]. Available: \url{https://arxiv.org/abs/2410.00371}
\BIBentrySTDinterwordspacing

\bibitem{dai2024racerrichlanguageguidedfailure}
\BIBentryALTinterwordspacing
Y.~Dai, J.~Lee, N.~Fazeli, and J.~Chai, ``Racer: Rich language-guided failure recovery policies for imitation learning,'' 2024. [Online]. Available: \url{https://arxiv.org/abs/2409.14674}
\BIBentrySTDinterwordspacing

\bibitem{cornelio2024recoverneurosymbolicframeworkfailure}
\BIBentryALTinterwordspacing
C.~Cornelio and M.~Diab, ``Recover: A neuro-symbolic framework for failure detection and recovery,'' 2024. [Online]. Available: \url{https://arxiv.org/abs/2404.00756}
\BIBentrySTDinterwordspacing

\bibitem{bougzime2025unlockingpotentialgenerativeai}
\BIBentryALTinterwordspacing
O.~Bougzime, S.~Jabbar, C.~Cruz, and F.~Demoly, ``Unlocking the potential of generative ai through neuro-symbolic architectures: Benefits and limitations,'' 2025. [Online]. Available: \url{https://arxiv.org/abs/2502.11269}
\BIBentrySTDinterwordspacing

\bibitem{zhang2024neurosymbolicaiexplainabilitychallenges}
\BIBentryALTinterwordspacing
X.~Zhang and V.~S. Sheng, ``Neuro-symbolic ai: Explainability, challenges, and future trends,'' 2024. [Online]. Available: \url{https://arxiv.org/abs/2411.04383}
\BIBentrySTDinterwordspacing

\bibitem{liu2023reflectsummarizingrobotexperiences}
\BIBentryALTinterwordspacing
Z.~Liu, A.~Bahety, and S.~Song, ``Reflect: Summarizing robot experiences for failure explanation and correction,'' 2023. [Online]. Available: \url{https://arxiv.org/abs/2306.15724}
\BIBentrySTDinterwordspacing

\bibitem{guo2024doremigroundinglanguagemodel}
\BIBentryALTinterwordspacing
Y.~Guo, Y.-J. Wang, L.~Zha, and J.~Chen, ``Doremi: Grounding language model by detecting and recovering from plan-execution misalignment,'' 2024. [Online]. Available: \url{https://arxiv.org/abs/2307.00329}
\BIBentrySTDinterwordspacing

\bibitem{wang2024largelanguagemodelsrobotics}
\BIBentryALTinterwordspacing
J.~Wang, Z.~Wu, Y.~Li, H.~Jiang, P.~Shu, E.~Shi, H.~Hu, C.~Ma, Y.~Liu, X.~Wang, Y.~Yao, X.~Liu, H.~Zhao, Z.~Liu, H.~Dai, L.~Zhao, B.~Ge, X.~Li, T.~Liu, and S.~Zhang, ``Large language models for robotics: Opportunities, challenges, and perspectives,'' 2024. [Online]. Available: \url{https://arxiv.org/abs/2401.04334}
\BIBentrySTDinterwordspacing

\bibitem{zhou2024codeasmonitorconstraintawarevisualprogramming}
\BIBentryALTinterwordspacing
E.~Zhou, Q.~Su, C.~Chi, Z.~Zhang, Z.~Wang, T.~Huang, L.~Sheng, and H.~Wang, ``Code-as-monitor: Constraint-aware visual programming for reactive and proactive robotic failure detection,'' 2024. [Online]. Available: \url{https://arxiv.org/abs/2412.04455}
\BIBentrySTDinterwordspacing

\bibitem{chen2024integrating}
X.~Chen, Y.~Cai, Y.~Mao, M.~Li, W.~Yang, W.~Xu, and J.~Wang, ``Integrating intent understanding and optimal behavior planning for behavior tree generation from human instructions,'' \emph{arXiv preprint arXiv:2405.07474}, 2024.

\bibitem{colledanchise2018behavior}
M.~Colledanchise and P.~{\"O}gren, \emph{Behavior trees in robotics and AI: An introduction}.\hskip 1em plus 0.5em minus 0.4em\relax CRC Press, 2018.

\bibitem{iovino2020survey}
M.~Iovino, E.~Scukins, J.~Styrud, P.~Ögren, and C.~Smith, ``A survey of behavior trees in robotics and ai,'' 2020.

\bibitem{styrud2024automaticbehaviortreeexpansion}
\BIBentryALTinterwordspacing
J.~Styrud, M.~Iovino, M.~Norrlöf, M.~Björkman, and C.~Smith, ``Automatic behavior tree expansion with llms for robotic manipulation,'' 2024. [Online]. Available: \url{https://arxiv.org/abs/2409.13356}
\BIBentrySTDinterwordspacing

\bibitem{8794128}
M.~Colledanchise, D.~Almeida, and P.~Ögren, ``Towards blended reactive planning and acting using behavior trees,'' in \emph{2019 International Conference on Robotics and Automation (ICRA)}, 2019, pp. 8839--8845.

\bibitem{wei2022chain}
J.~Wei, X.~Wang, D.~Schuurmans, M.~Bosma, E.~H. Chi, Q.~V. Le, and D.~Zhou, ``Chain of thought prompting elicits reasoning in large language models,'' \emph{Advances in neural information processing systems}, 2022.

\bibitem{liu2024groundingdinomarryingdino}
\BIBentryALTinterwordspacing
S.~Liu, Z.~Zeng, T.~Ren, F.~Li, H.~Zhang, J.~Yang, Q.~Jiang, C.~Li, J.~Yang, H.~Su, J.~Zhu, and L.~Zhang, ``Grounding dino: Marrying dino with grounded pre-training for open-set object detection,'' 2024. [Online]. Available: \url{https://arxiv.org/abs/2303.05499}
\BIBentrySTDinterwordspacing

\bibitem{ravi2024sam2segmentimages}
\BIBentryALTinterwordspacing
N.~Ravi, V.~Gabeur, Y.-T. Hu, R.~Hu, C.~Ryali, T.~Ma, H.~Khedr, R.~Rädle, C.~Rolland, L.~Gustafson, E.~Mintun, J.~Pan, K.~V. Alwala, N.~Carion, C.-Y. Wu, R.~Girshick, P.~Dollár, and C.~Feichtenhofer, ``Sam 2: Segment anything in images and videos,'' 2024. [Online]. Available: \url{https://arxiv.org/abs/2408.00714}
\BIBentrySTDinterwordspacing

\end{thebibliography}
\bibliographystyle{bib/IEEEtran}

\end{document}